%% file: colm2026_conference.tex
\documentclass{article} 

\usepackage[preprint]{colm2026_conference}

\usepackage{microtype}
\usepackage{graphicx}     
\usepackage{subcaption}   
\usepackage{wrapfig}
\usepackage{caption}
\usepackage{booktabs}     
\usepackage{tcolorbox}
\usepackage[utf8]{inputenc}
\usepackage{textcomp}

\input{math_commands.tex}

\usepackage{hyperref}
\usepackage{url}

\usepackage{lineno}

\definecolor{darkblue}{rgb}{0, 0, 0.5}
\hypersetup{colorlinks=true, citecolor=darkblue, linkcolor=darkblue, urlcolor=darkblue}

\newcommand{\methodname}{\ensuremath{\text{S}^{3}\text{-R1}}}
\title{\methodname: Learning to Retrieve and Answer Step-by-Step with Synthetic Data}

\author{Harsh Goel\thanks{Work done during an internship at Google DeepMind.}\\
The University of Texas at Austin\\
\And
Akhil Udathu \\
Google DeepMind \\
\And
Susmija Jabireddy \\
Google DeepMind \\
\And
Pradnesh Kalkar\\
Google DeepMind \\
\And
Atharva Parulekar \\
Google DeepMind \\
}

\begin{document}

\ifcolmsubmission
\linenumbers
\fi

\maketitle

\input{sections/abstract}

\section{Introduction}
\label{sec:intro}

\input{sections/intro}

\section{Related Works}
\label{sec:related}
\input{sections/relatedworks}

\section{Preliminaries}
\label{sec:prelim}
\input{sections/preliminaries}

\section{Method}
\label{sec:method}
\input{sections/method}
\label{sec:results}
\input{sections/results}

\section{Conclusion}
\label{sec:conclusion}
\input{sections/conclusion}




\bibliography{colm2026_conference}
\bibliographystyle{colm2026_conference}

\appendix
\input{sections/appendix}
\end{document}

%% file: math_commands.tex
\usepackage{amsmath,amsfonts,bm}

\def\eqref#1{equation~\ref{#1}}

\def\1{\bm{1}}

\DeclareMathAlphabet{\mathsfit}{\encodingdefault}{\sfdefault}{m}{sl}
\SetMathAlphabet{\mathsfit}{bold}{\encodingdefault}{\sfdefault}{bx}{n}



%% file: sections/abstract.tex
\begin{abstract}

Reinforcement learning (RL) post-training has enabled newer capabilities in models, such as agentic tool-use for search. However, these models struggle primarily due to limitations with sparse outcome-based rewards and a lack of training data that encapsulates questions of differing hardness, which results in models not performing deeper searches with tools to collect evidence for question-answering. To address these limitations, we introduce \methodname (Synthetic data and stabilized Search R1), a framework that couples a data-centric approach with denser learning signals. We first develop a synthetic generation and curation pipeline that programmatically derives diverse, multi-hop questions from existing documents. This pipeline incorporates a retrieval-based verification step to specifically isolate questions of intermediate difficulty. We then pair this expanded training set with a reward structure that evaluates both intermediate search quality and the correctness of the final answer. This setup directly mitigates the credit assignment problems inherent to sparse rewards. Our evaluations show that \methodname\ outperforms existing baselines by learning more effective search and synthesis strategies, yielding up to a 10\% improvement in robust generalization on out-of-domain datasets.
\end{abstract}

%% file: sections/intro.tex
Large Language Models (LLMs) have demonstrated remarkable capabilities across multiple domains, driving a rapid transition toward an agentic paradigm where models act as autonomous problem-solvers \citep{gemini2024, anthropic2024, openai2024}. However, their reliance on static, parametric knowledge limits their effectiveness on tasks requiring access to real-time information, specifically in Question-Answering (QA). While Retrieval-Augmented Generation (RAG) \citep{lewis2020retrieval} alleviates this limitation, contemporary LLMs still heavily struggle with complex, multi-hop questions that require iterative reasoning and strategic tool-use to retrieve the correct content \citep{trivedi2022musique, trivedi2023interleaving}.

To orchestrate this sequential tool-use, recent methods frame multi-hop QA as an interactive process, utilizing external search engines to construct reasoning trajectories \citep{jin2025searchr1, song2025r1searcher}. Consequently, Reinforcement Learning (RL) has emerged as a powerful paradigm for training these agentic policies \citep{guo2025deepseekr1}. Frameworks like Search-R1 \citep{jin2025searchr1} and R1-Searcher \citep{song2025r1searcher} successfully apply RL to teach models to autonomously generate search queries, process retrieved results, and perform step-by-step reasoning. These methods typically optimize the model using outcome-based rewards derived solely from the final answer's correctness.

While RL-based post-training has improved QA, gains are increasingly constrained by the training distribution and sparse, outcome-only rewards. If training data rarely requires multi-hop reasoning and rewards score only the final answer, RL overfits to short-horizon patterns and fails to generalize across hops. Existing synthetic QA pipelines mostly generate reasoning traces for a fixed set of questions \citep{goldie2025synthetic}, which helps credit assignment but does little to broaden the question distribution itself. By contrast, self-play and synthetic generation in formal domains (e.g., math and programming) explicitly expand problem difficulty and diversity \citep{chen2025self,zeng2025rlve}.

Inspired by these formal domain approaches that systematically expand the training distribution across varying difficulty levels, we introduce \methodname\ that first aims to construct a similar paradigm for complex, multi-hop QA. Our \textbf{key technical insight} is to synthesize verified yet solvable question instances by utilizing a frontier model to mutate problems that the base agent fails to solve. To ensure these questions are viable for training, we implement a verification pipeline that filters for both factual grounding and empirical retrieval difficulty. By confirming that a question is answerable given perfect information and remains solvable within a noisy, high-recall retrieval environment, we generate an intermediate-difficulty band that is challenging yet learnable. To address the bottlenecks imposed by sparse, outcome-only supervision, we pair this data with a retrieval-aware reward that explicitly incentivizes search quality and evidence selection rather than only final-answer correctness. Through extensive experiments, \methodname\ substantially improves over strong baselines on multiple multi-hop QA benchmarks.


%% file: sections/relatedworks.tex
\subsection{Large Language Models for Retrieval, Search, and Question Answering}

While Large Language Models (LLMs) exhibit strong reasoning capabilities~\citep{guo2025deepseekr1}, their reliance on static, parametric knowledge makes them susceptible to hallucinations and knowledge cutoffs~\citep{zhang2025siren}. Retrieval-Augmented Generation (RAG) mitigates this by grounding LLM outputs in external evidence~\citep{lewis2020retrieval, gao2023retrieval}. However, the standard ``retrieve-then-read'' pipeline is brittle: retrieved context can include distracting or stale information (``context rot'') ~\citep{jin2024longcontext}) and document ordering can be suboptimal~\citep{chang2025main}. As a result, early work focused on improving retrieval quality, for example via query rewriting~\citep{wilson2025contextualizing,wang2025maferw,kong2024prewrite}.
A more powerful paradigm is active tool use, where the LLM acts as an agent that iteratively reasons, searches, and updates its evidence. Early approaches such as Chain-of-Thought (CoT)~\citep{wei2022chainofthought}, ReAct~\citep{yao2023react}, and IRCoT~\citep{trivedi2023interleaving} relied on carefully designed prompting frameworks for reasoning and tool use, while supervised fine-tuning (SFT) requires expensive, large-scale labeled trajectories~\citep{schick2023toolformer}. However,  for complex multi-hop questions, ``semantic drift " ~\citep {mavi2024multi} from imprecise search queries during tool use often results in degraded performance. Therefore, more recent systems, such as Search-R1 and R1-Searcher, instead leverage reinforcement learning to train search-and-reason policies that decide what to query, how to use retrieved evidence, and when to stop to answer questions.

\subsection{Reinforcement Learning for Large Language Models}
While Proximal Policy Optimization (PPO) \citep{schulman2017proximal} established the foundation for aligning large language models \citep{ouyang2022training, kaelbling1996reinforcement, sutton1999reinforcement}, the overhead induced by learning a value model has driven the search for simpler alternatives. Direct optimization strategies like DPO \citep{rafailov2023direct} and SimPO \citep{meng2024simpo} bypass the reward model entirely, but frequently encounter off-policy degradation \citep{pang2024iterative, hsu2024grounding}. Conversely, recent on-policy methods focus on streamlining the RL pipeline itself. Algorithms such as GRPO \citep{shao2024deepseekmath}, DAPO \citep{yu2025dapo}, RLOO \citep{ahmadian2024back}, and GSPO \citep{zheng2025group} maintain training stability while completely discarding the value network. Because these critic-free frameworks excel at eliciting complex reasoning from sparse outcome rewards \citep{guo2025deepseekr1}, we utilize them in \methodname\ to post-train LLMs for multi-hop search following prior works \citep{jin2025searchr1,song2025r1searcher}.

\subsection{Synthetic Data for Search and Tool Use}
Training on synthetic data has shown to improve many LLMs ~\citep{nadas2025synthetic}.
Many synthetic data generation methods for reasoning, such as STaR~\citep{zelikman2022star}, Rejection Finetuning (RFT)~\citep{yuan2023scalingrelationshiplearningmathematical}, ReST~\citep{gulcehre2023reinforced}, and ReSTEM~\citep{singh2023beyond}, rely on a 'generate-and-filter' approach. These techniques prompt a model to produce reasoning traces, such as chains-of-thought, and then perform Supervised Fine-Tuning (SFT) exclusively on the responses that lead to a correct final answer. In the retrieval and QA domain, LERET ~\citep{hsu2024grounding} relies on generating synthetic instances of intermediate queries seeded from in-context examples are trained via DPO on outcomes. More recently, ~\citep{goldie2025synthetic} showed that learning from synthetic reasoning steps from a strong teacher can improve tool use in multi-turn settings. In contrast, our paper uses a stronger model to generate synthetic intermediate-difficulty questions from cases where the base agent is weak, enabling the weaker agent to learn from novel but easier problem instances.

%% file: sections/preliminaries.tex
\subsection{Multi-hop Search}
\label{sec:prelim_multihop}
Many methods, such as Search-R1 \citep{jin2025searchr1} and R1-Searcher \citep{song2025r1searcher} use a structured rollout for the LLM to search during QA. The generation of a complete response trajectory $y$, denoted as $y \sim \pi_{\theta}(\cdot | x, \mathcal{R})$ for a given prompt $x$ and a search engine $\mathcal{R}$, is an interleaved sequence of text generation and tool calls, governed by a token-based protocol. The process begins with the LLM generating thoughts within the \texttt{<think>} and \texttt{</think>}. When external information is required, the model searches with tokens, \texttt{<search>} and \texttt{</search>}. The agent executes the search query through a search engine, and the retrieved results are enclosed within \texttt{<information>} and \texttt{</information>} tokens, and added to the context for further reasoning and generation. This trajectory is terminated when the model produces a final answer within \texttt{<answer>} and \texttt{</answer>} tokens or when a preset limit on search calls is reached.

\subsection{Group Relative Policy Optimization for Multi-hop Search}
\label{sec:prelim_grpo}
Several RL methods have been introduced to post-train LLMs to optimize a reward function. Proximal Policy Optimization (PPO) \citep{schulman2017proximal} optimizes the policy using a learned value function, whereas Group Relative Policy Optimization (GRPO), proposed by Shao et al. (2024), avoids the need for an explicit value function.

GRPO establishes a baseline for policy updates using the average reward from a group of sampled trajectories. For each input prompt $x$, GRPO \citep{shao2024deepseekmath} samples a group of $G$ responses, $\{y_1, y_2, \dots, y_G\}$, from a policy  $\pi_{\text{old}}$. The current policy, $\pi_{\theta}$, is then optimized by maximizing the following objective function:
\begin{equation}
\label{eq:grpo}
\begin{aligned}
J_{\text{GRPO}}(\theta) = \mathbb{E}_{x \sim \mathcal{D}, \{y_i\}_{i=1}^G \sim \pi_{\text{old}}} \Bigg[
    \frac{1}{G} \sum_{i=1}^{G} \frac{1}{\sum_{t=1}^{|y_i|} I(y_i, t)} \sum_{t=1, I(y_i, t)=1}^{|y_i|} \min \Bigg( 
    \quad \frac{\pi_{\theta}(y_{i,t} | x, y_{i,<t})}{\pi_{\text{old}}(y_{i,t} | x, y_{i,<t})} \hat{A}_{i,t}, \\
    \quad \text{clip}\left(\frac{\pi_{\theta}(y_{i,t} | x, y_{i,<t})}{\pi_{\text{old}}(y_{i,t} | x, y_{i,<t})}, 1-\epsilon, 1+\epsilon\right) \hat{A}_{i,t}
    - \beta D_{\text{KL}}[\pi_{\theta}(y_{i,t} | x, y_{i,<t}) || \pi_{\text{ref}}(y_{i,t} | x, y_{i,<t})] \Bigg)
\Bigg]
\end{aligned}
\end{equation}

Here, $\epsilon$ and $\beta$ are hyperparameters controlling the clipping ratio and the strength of the KL regularization, respectively, and $\pi_{\text{ref}}$ is the reference policy. The advantage, $\hat{A}_{i,t} = \frac{r_i - mean(r)}{std(r)}$, where $r$ denotes the per-sample reward computed within each group.

GRPO directly adds the KL divergence between the trained policy $\pi_{\theta}$ and the reference policy $\pi_{\text{ref}}$ to the loss function as a regularization term. For search, a token mask $I(y_i, t)$  is applied to mask the information obtained from external sources \citep{jin2025searchr1}.

%% file: sections/method.tex
This section outlines our training methodology for \methodname. To successfully train models for complex, multi-hop reasoning, we pivot away from relying solely on standard optimization tweaks and instead emphasize a data-centric approach coupled with dense reward signals. We first broaden the training distribution via a rigorous synthetic question generation pipeline that mines hard seeds, generates diverse questions, and filters them based on retrieval-solvability. We then pair this enhanced data distribution with a retrieval-aware reward that augments outcome-only supervision with a recall-based signal to directly reward search quality, alongside final answer correctness.

\input{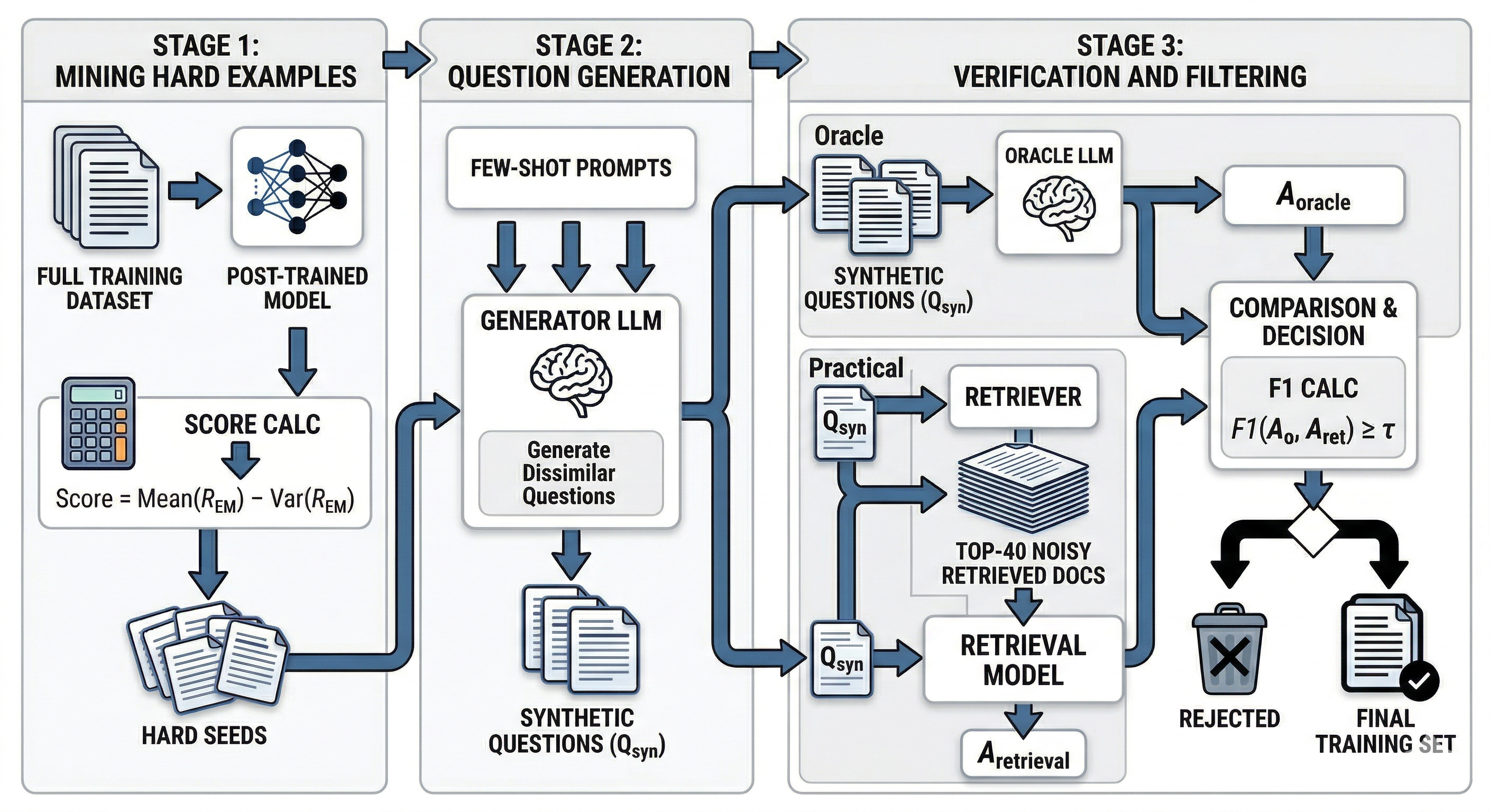}

\subsection{Synthetic Data Generation for Enhanced Training}

To broaden the coverage of extended reasoning and rebalance the training distribution, we augment our training set with synthetically generated questions via a pipeline comprising three distinct phases: hard example mining (isolating hard questions), question generation (converting hard questions to synthetic questions), and retrieval-based verification (finding questions of intermediate difficulty with their answers).

\subsubsection{Identifying Hard Anchor Instances}

We synthesize questions by first identifying hard anchor questions that are typically characterized by lower average accuracies and high stochasticity from the model. We begin by post-training the model on the entire existing dataset $\mathcal{Q}$. For each prompt $q \in \mathcal{Q}$, we evaluate its empirical difficulty using the best-performing checkpoint by sampling $K=5$ independent reasoning trajectories. We then calculate a pessimistic ~\citep{la2013actor} solvability score $Score(q)$ based on the mean $\mathbb{E}[R_{\text{F1}}]$ and sample variance $\text{Var}[R_{\text{F1}}]$ of the F1 scores as  
$$Score(q) = \mathbb{E}_{a \sim \pi_{\theta}}[R_{\text{F1}}(q,a)] - \text{Var}_{a \sim \pi_{\theta}}[R_{\text{F1}}(q,a)].$$

Because the F1 score provides a continuous evaluation signal, incorporating the variance penalty per group establishes a lower confidence score ~\citep{buckman2020importance} that explicitly heavily penalizes questions exhibiting high epistemic uncertainty. This fundamentally reorders questions, ensuring that unstable, high-variance instances are driven further down the lower tail alongside instances of consistent failure. Consequently, this scoring mechanism isolates the actual boundary of the model's competence, implicitly identifying multi-hop questions as hard anchor examples. We then select the 10,000 questions with the lowest $Score(q)$ values as anchor examples for synthetic question generation.

\subsubsection{Dissimilarity-Driven Question Generation}

In the second phase, we utilize a few-shot prompting strategy with a highly capable generator model (Gemini 2.5 Pro) to synthesize high-quality, relevant questions. We construct a context-rich prompt comprising several in-context examples, where each exemplar pairs a set of source documents with a corresponding high-quality question sampled from our pool of mined anchor instances. Then, to this context, we append the target document set along with its original anchor question to generate a new synthetic question grounded in the target document set. Crucially, we instruct the generator to synthesize questions that diverge from the original anchor question and further filter out highly similar questions to the original anchor question. Moreover, by systematically randomizing the in-context exemplars, the generator model generates a more varied distribution of new problems. The exact prompts utilized for this generation process are detailed in the Appendix.

\subsubsection{Verification and Filtering}

A generated question is only viable for training if it is both factually solvable from the source material and practically retrievable given the limitations of the standard retrieval tool utilized. To ensure these conditions are met, we introduce a two-phase verification process for the synthetic questions. First, we establish an upper bound on factual solvability by prompting the generator model (Gemini 2.5 Pro) alongside the ground-truth documents to produce an oracle answer, $A_{\text{oracle}}$. Second, we evaluate the empirical retrieval difficulty. Because the downstream model interacts with the corpus using a relatively weak lexical retriever (BM25), questions that demand excessively complex semantic matching to surface the relevant text are practically unsolvable during training. To filter out these intractable instances, we fetch the top-k documents via BM25 and tasking the generator to produce a retrieval-based answer, $A_{\text{retrieval}}$, relying strictly on this retrieved context.

We set $k=40$ to create a high-recall, low-precision environment; if the generator can extract the necessary evidence from this noisy context, it confirms that the weak retriever is at least capable of surfacing the required information. During actual RL training, the agent is heavily constrained, permitted to retrieve only 5 documents per turn across a maximum of 5 turns. Consequently, questions that pass this 40-document verification step represent an optimal challenge: they are difficult enough to incentivize the model to learn highly precise, iterative search strategies, yet grounded enough to be solvable without requiring a prohibitively powerful semantic retriever.

Finally, we evaluate the agreement between the two answers by calculating a token-level F1 score, retaining only questions where $F1(A_{\text{oracle}}, A_{\text{retrieval}}) \geq \tau$, with $\tau$ serving as the acceptance threshold. We prioritize F1 over EM to prevent aggressively discarding questions. This verification pipeline yields a filtered synthetic dataset that is challenging, factually grounded, and reliably solvable under the constraints of a multi-hop retrieval environment.

\subsection{Reward Formulation}

Standard reward functions for question-answering are typically sparse, relying entirely on a binary Exact Match (EM) signal at the end of a trajectory. For multi-hop QA, this sparse formulation suffers from severe credit assignment issues; a model might execute three perfect search queries but fail on the final synthesis, receiving a zero reward that inadvertently penalizes its excellent search behavior. To provide a denser, step-aware learning signal, we formulate a composite reward that evaluates both the final correctness and the intermediate quality of the information retrieval process. Here, $R$ averages the Exact Match score and the Cumulative Recall of the retrieved documents:$$R = \frac{R_{\text{EM}} + R_{\text{Recall}}}{2}.$$
Here, $R_{\text{EM}} \in \{0, 1\}$ is the binary correctness of the final answer, and $R_{\text{Recall}}$ calculates the fraction of ground-truth documents successfully retrieved across all search turns in the trajectory. By rewarding intermediate retrieval, the policy receives a positive learning signal for searching documents even if the final answer is incorrect.

\subsection{Stabilization of Reinforcement Learning Training}

Because standard GRPO can be susceptible to premature convergence and erratic updates over long multi-hop rollouts, we integrate three established stabilization techniques to ensure robust learning. First, following DAPO \citep{yu2025dapo}, we apply double clipping with an elevated upper bound to preserve learning signals for low-probability exploration tokens. Concurrently, we increase the entropy loss coefficient \citep{schulman2017proximal} to actively prevent the policy from collapsing into suboptimal, deterministic search patterns early in training. Finally, to address gradient instability caused by negative advantages \citep{jin2025searchr1}, we follow \cite{ye2020mastering} by applying a strict lower bound to the importance ratio $r_t(\theta)$. Specifically, when the advantage $\hat{A}_t < 0$, we cap the penalty magnitude by lower-bounding the standard GRPO objective with $\epsilon_{\text{neg}} \hat{A}_t$ (where $\epsilon_{\text{neg}} > 1$).

%% file: figures/pipeline.tex
\begin{figure}[ht]
    \centering 
    \includegraphics[width=0.8\linewidth]{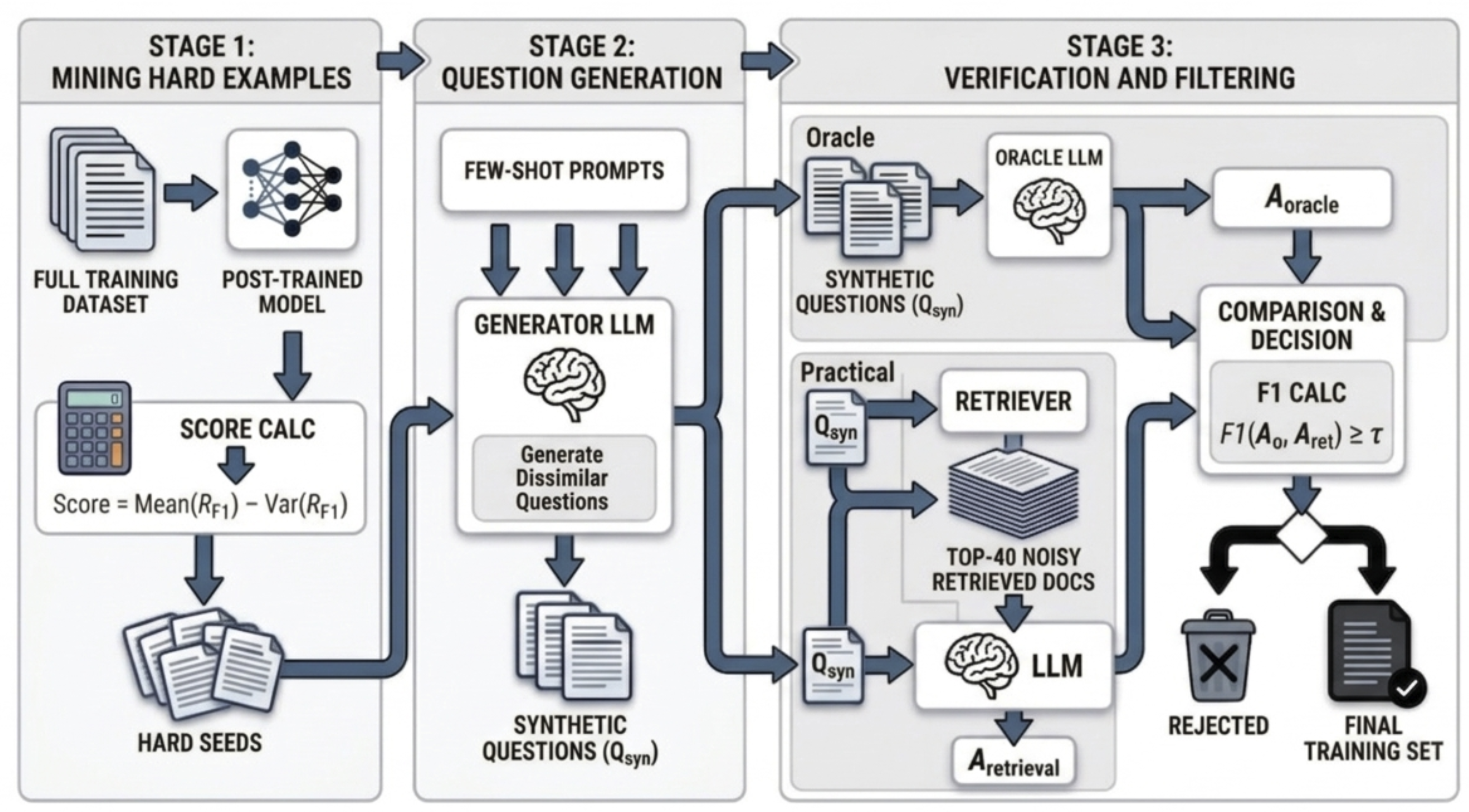}
    \caption{\textbf{Synthetic Data Generation Pipeline.} We mine hard anchor questions by scoring training prompts with a risk-adjusted solvability metric (mean minus variance over 5 rollouts) and selecting the lowest-scoring 10K instances. Conditioned on each anchor’s evidence documents, a generator model (Gemini 2.5 Pro) produces dissimilar synthetic questions. We then verify solvability under retrieval by comparing an oracle answer from the ground-truth documents to an answer produced from BM25 top-40 retrieved documents, retaining questions whose token-level F1 exceeds a threshold.}
    \label{fig:pipeline} 
    \vspace{-1em}
\end{figure}

%% file: sections/results.tex
\section{Experiments}
\input{plots/overall_performance}

We conduct a series of experiments to comprehensively evaluate the effectiveness of our proposed methodology, which combines RL post-training enhancements with a synthetic data pipeline. Our evaluation is designed to answer the following key research questions:
\begin{enumerate}
    \item \textbf{Impact of RL Enhancements:} To what extent do the RL post-training enhancements improve performance over the baseline Search-R1?
    \item \textbf{Impact of Synthetic Data:} How does augmenting the training data with our synthetic, hard examples affect both in-domain and out-of-domain performance?
    \item \textbf{Ablation of Synthetic Data:} What is the overall contribution of the different components of our synthetic data generation pipeline?
\end{enumerate}

\paragraph{Datasets and Metrics.}
We evaluate our models on four challenging multi-hop question-answering benchmarks: Musique~\citep{trivedi2022musique}, HotpotQA~\citep{yang2018hotpotqa}, 2WikiMultiHopQA~\citep{ho2020constructing}, and CofCA~\citep{wu2024cofca}. All models are trained exclusively on the Musique training set. We evaluate performance using two primary metrics. \textbf{Exact Match (EM)} measures the fraction of model-generated answers that exactly match the ground truth. \textbf{Recall} calculates the fraction of necessary ground-truth documents that were successfully retrieved during the model's search steps.
\input{plots/perfromance_average}

\paragraph{Baselines and Models.}
Our primary experiments are conducted using the open-source \textbf{Qwen2.5-7B Instruct} model for its strong instruction-following \citep{jin2025searchr1}. We compare against a spectrum of baselines to provide a thorough analysis:
\begin{itemize}
    \item \textbf{Direct:} The model is asked to answer the question based on its internal knowledge. 
    \item \textbf{RAG Methods:} Standard Retrieval-Augmented Generation (RAG) with context obtained through a BM25 retriever; Chain-of-Thought with RAG, where the model is asked to think carefully step by step to answer questions with the provided context (CoT + RAG); Multi-hop RAG baseline similar to the setup in \ref{sec:prelim_multihop}.
    \item \textbf{Trained Models:} We post-train Qwen2.5-7b Instruct with Search-R1 and \methodname.
\end{itemize}

\input{figures/rl_ablation}
We also evaluate \textbf{\methodname} against a suite of advanced RAG prompting strategies with \textbf{Gemini 2.5 Pro}, including standard RAG, CoT + RAG, a decomposition-based approach (Decomp + RAG), which first decomposes the original question to sub-questions and retrieves documents for each sub-question before answering the original question, and Multi-Hop RAG.

\paragraph{Experimental Setup.}
All models are trained solely on the Musique dataset to rigorously assess zero-shot generalization performance on HotpotQA, 2WikiMultiHopQA, and CofCA. We use a consistent retriever (BM25) to ensure fair comparisons. While stronger retrievers would improve results \citep{jin2025empirical}, we expect this to improve all methods proportionally. Further details regarding hyperparameters, the training process, and our experimental infrastructure can be found in the Appendix.

\section{Discussions}

\subsection{Impact of our RL Algorithm Enhancements}

To isolate the effect of our proposed RL algorithm enhancements, we first compare the performance of our model (\textbf{\methodname\ - No Synthetic}) against the baseline \textbf{Search-R1}. As shown in Table~\ref{tab:multihop_qa_performance}, our method demonstrates a notable improvement in both Exact Match and Recall across all datasets. This indicates that the algorithmic modifications, including negative advantage clipping and the denser reward function, stabilize training.

Furthermore, we analyze the training dynamics of both models. Figure~\ref{fig:rl_ablation} plots the reward over the course of training and reveals a critical difference in stability. While the baseline Search-R1 is prone to performance collapses and high variance, our enhanced algorithm maintains a stable and monotonically increasing reward curve for a much longer training duration. This stability is particularly beneficial for complex queries that require longer reasoning chains or multiple search "hops." By mitigating erratic policy updates, the model can learn to make longer learning progress on harder problems, leading to superior performance on the most challenging questions.

\subsection{Impact of Synthetic Data}

We assess the contribution of our synthetic data pipeline by comparing the full model (\textbf{\methodname}) to a version trained only on the original data. The results in Table~\ref{tab:multihop_qa_performance}, Table~\ref{tab:multihop_qa_performance_avg}, and Table~\ref{tab:cofca_musique_per_hop} show that the inclusion of synthetic data yields a significant additional performance boost. Importantly, these improvements extend beyond the in-domain Musique benchmark to out-of-domain datasets, indicating that our augmentation enhances generalization rather than overfitting to a single corpus.

The training curves in Figure~\ref{fig:rl_ablation} further corroborate the influence of the synthetic questions. The model trained on the mixture of synthetic and original questions consistently achieves a higher average  pass@1 performance averaged over 8 passes. We believe that by injecting synthetic questions of intermediate difficulty, the model develops transferable multi-hop reasoning ability. 

\input{plots/hop_performance}
\input{figures/synthetic_ablation}
\subsection{Ablation on Synthetic Data Generation}

The effectiveness of our synthetic data pipeline depends on two key hypotheses: (1) that seeding from "hard" examples is crucial, and (2) that a rigorous verification process is necessary to ensure data quality. To validate these design choices, we conduct an ablation study with two alternative training runs:
\begin{itemize}
    \item \textbf{Random Verified:} A model trained with synthetic data generated from randomly selected, easier examples instead of the hard-mined seeds.
    \item \textbf{Hard Unverified:} A model trained with synthetic data generated from hard-mined seeds but without the final verification and filtering step.
    \item \textbf{Only Synthetic, Hard Verified:}  A model trained only on synthetic data generated from hard-mined seeds with verification. 
\end{itemize}

Figure~\ref{fig:synth_ablation_combined} presents the average pass@1 performance over recall and exact match for these ablations compared to our full method. The results clearly demonstrate that both components are critical. The model trained on unverified data performs worse, indicating that the verification step is essential for filtering out noisy or unsolvable questions that would otherwise introduce misleading signals into the training process. This is further corroborated by the performance of the model on only synthetic data, where we observe no learning progress. Moreover, when we replace mined hard seeds with randomly initialized seeds, training becomes unstable after roughly 300 steps, and therefore shows a diminished performance gain. These results confirm that our hard-example mining strategy successfully targets the model's weaknesses and drives more meaningful learning progress.

%% file: plots/overall_performance.tex
\begin{table}[ht]
\centering
\caption{Pass@8 Performance Comparison on Multi-Hop QA Datasets}
\label{tab:multihop_qa_performance}
\resizebox{\textwidth}{!}{%
\begin{tabular}{@{}lcccccccc@{}}
\toprule
 & \multicolumn{2}{c}{\textbf{CofCA}} & \multicolumn{2}{c}{\textbf{Musique}} & \multicolumn{2}{c}{\textbf{2Wiki}} & \multicolumn{2}{c}{\textbf{HotpotQA}} \\
\cmidrule(lr){2-3} \cmidrule(lr){4-5} \cmidrule(lr){6-7} \cmidrule(lr){8-9}
\textbf{Methods} & \textbf{EM} & \textbf{Recall} & \textbf{EM} & \textbf{Recall} & \textbf{EM} & \textbf{Recall} & \textbf{EM} & \textbf{Recall} \\
\midrule
\multicolumn{9}{@{}l}{\textbf{Qwen2.5-7b Instruct}} \\
Direct Inference & 0.288 & - & 0.039 & - & 0.378 & - & 0.220 &  - \\
RAG & 0.283 & 53.2 & 0.028 & 43.5 & 0.205 & 62.3 & 0.228 & 72.5 \\
CoT + RAG & 0.403 & 53.2 & 0.060 & 43.5 & 0.299 & 62.3 & 0.328 & 72.5 \\
Multihop + RAG &  0.335 & 53.2 & 0.098 & 61.1 & 0.400 & 83.0 & 0.354 & 80.7 \\
Search-R1 & 0.401 & 52.7 &0.408& 69.9 & 0.586 & 90.4 & 0.513 & 81.3 \\
\textbf{\methodname - No Synthetic} & 0.521 & 54.0 & 0.448 & 69.9 & 0.663 & 88.2 & 0.576 & 83.3 \\
\textbf{\methodname} & \textbf{0.605} & 52.9 & \textbf{0.460} & 69.1 & \textbf{0.694} & \textbf{90.3} & \textbf{0.599} & \textbf{85.5} \\
\midrule
\multicolumn{9}{@{}l}{\textbf{Gemini 2.5 Pro}} \\
RAG & 0.418 & 43.3 & 0.171 & 50.9 & 0.325 & 66.9 & 0.352 & 63.8 \\
COT + RAG & 0.514 & 55.2 & 0.144 & 51.1 & 0.297 & 67.5 & 0.385 & 78.7 \\
Decomp + RAG & 0.522 & 55.3 & 0.199 & 62.6 & 0.329 & 70.7 & 0.430 & 83.0 \\
MultiHop + RAG & 0.573 & \textbf{55.7} & 0.452 & \textbf{81.5} & 0.651 & 87.8 & 0.587 & 85.1 \\
\bottomrule
\end{tabular}%
}
\end{table}

%% file: plots/perfromance_average.tex
\begin{table}[ht]
\centering
\caption{Pass@1 (8 sample average) Performance Comparison on Multi-Hop QA Datasets.}
\label{tab:multihop_qa_performance_avg}
\resizebox{\textwidth}{!}{%
\begin{tabular}{@{}lcccccccc@{}}
\toprule
 & \multicolumn{2}{c}{\textbf{Cofca}} & \multicolumn{2}{c}{\textbf{Musique}} & \multicolumn{2}{c}{\textbf{2Wiki}} & \multicolumn{2}{c}{\textbf{HotpotQA}} \\
\cmidrule(lr){2-3} \cmidrule(lr){4-5} \cmidrule(lr){6-7} \cmidrule(lr){8-9}
\textbf{Methods} & \textbf{EM} & \textbf{Recall} & \textbf{EM} & \textbf{Recall} & \textbf{EM} & \textbf{Recall} & \textbf{EM} & \textbf{Recall} \\
\midrule
\multicolumn{9}{@{}l}{\textbf{Qwen2.5-7b Instruct}} \\
Direct Inference & 0.057 & - & 0.007 & - & 0.091& - & 0.052 &  - \\
RAG & 0.047 & 53.2 & 0.005 & 43.5 & 0.040 & 62.3 & 0.044 & 72.5 \\
CoT + RAG & 0.092 & 53.2 & 0.011 & 43.5 & 0.068 & 62.3 & 0.082 & 72.5 \\
Multihop + RAG &  0.068 & 29.2 & 0.015 & 24.3 & 0.083 & 41.1 & 0.079 & 36.5 \\
Search-R1 & 0.184 & 38.2 &0.254 & 53.8 & 0.338 & 73.1 & 0.298 & 56.3 \\
\textbf{\methodname - No Synthetic} & 0.358 & 46.4 & 0.317 & 60.5 & 0.475 & 77.7 & 0.441 & 72.1 \\
\textbf{\methodname} & \textbf{0.455} & \textbf{49.7} & \textbf{0.345} & \textbf{61.9} & \textbf{0.519} & \textbf{83.8} & \textbf{0.483} & \textbf{78.1} \\

\bottomrule
\end{tabular}%
}
\vspace{-2em}
\end{table}

%% file: figures/rl_ablation.tex
\begin{wrapfigure} [17]{r}{0.5\textwidth}
    \centering
    \includegraphics[width=0.8\linewidth]{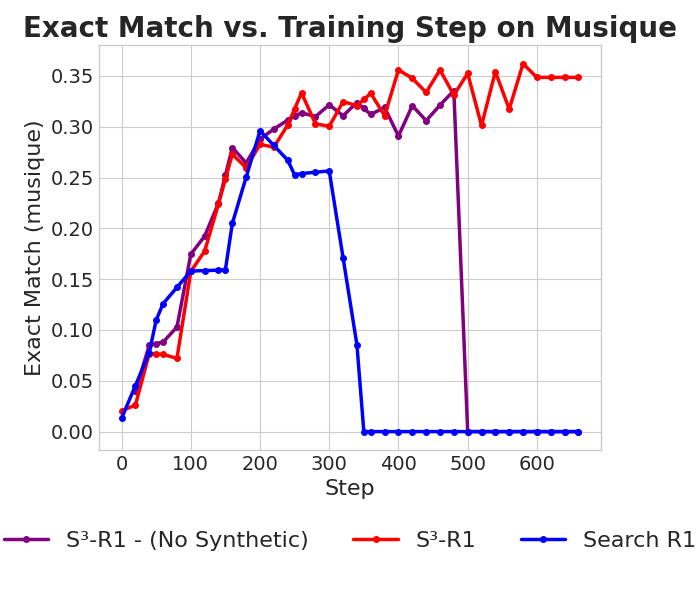}
      \captionsetup{width=0.\textwidth, justification=raggedright, singlelinecheck=false}
      \vspace{-1em}
    \caption{\textbf{Impact of RL algorithm changes on training.} We show that post-training Qwen2.5-7B without RL enhancements (purple) is more stable than Search-R1 (blue). }
    \label{fig:rl_ablation}
\end{wrapfigure}

%% file: plots/hop_performance.tex
\begin{table}[ht]
\centering
\caption{Per-Hop Exact Match (EM) Pass@8 Performance on Cofca and Musique Datasets}
\label{tab:cofca_musique_per_hop}
\resizebox{\textwidth}{!}{%
\begin{tabular}{@{}lcccccccc@{}}
\toprule
 & \multicolumn{4}{c}{\textbf{Cofca Exact Match Score}} & \multicolumn{4}{c}{\textbf{Musique Exact Match Score}} \\
\cmidrule(lr){2-5} \cmidrule(lr){6-9}
\textbf{Methods} & \textbf{2-Hop} & \textbf{3-Hop} & \textbf{4-Hop} & \textbf{Overall EM} & \textbf{2-Hop} & \textbf{3-Hop} & \textbf{4-Hop} & \textbf{Overall EM} \\
\midrule
\multicolumn{9}{@{}l}{\textbf{Qwen2.5-7b Instruct}} \\
Multihop + RAG & 0.275 & 0.432 & 0.300 & 0.335 & 0.143 & 0.060 & 0.033 & 0.098 \\
Search-R1 & 0.406 & 0.329 & 0.462 & 0.401 & 0.501 & 0.347 & 0.231 & 0.408 \\
\textbf{\methodname - No Synthetic} & 0.437 & 0.558 & 0.570 & 0.521 & \textbf{0.538} & 0.361 & 0.337 & 0.448 \\
\textbf{\methodname} & \textbf{0.482} & 0.653 & \textbf{0.679} & \textbf{0.605} & 0.526 & 0.409 & \textbf{0.351} & \textbf{0.460} \\
\midrule
\multicolumn{9}{@{}l}{\textbf{Gemini 2.5 Pro}} \\
RAG & 0.413 & 0.457 & 0.384 & 0.418 & 0.218 & 0.136 & 0.088 & 0.171 \\
COT + RAG & 0.390 & 0.558 & 0.592 & 0.514 & 0.198 & 0.100 & 0.049 & 0.144 \\
Decomp + RAG & 0.390 & 0.555 & 0.618 & 0.522 & 0.255 & 0.160 & 0.097 & 0.199 \\
MultiHop + RAG & 0.460 & \textbf{0.672} & 0.589 & 0.573 & 0.529 & \textbf{0.426} & 0.262 & 0.452 \\
\bottomrule
\end{tabular}%
}
\end{table}

%% file: figures/synthetic_ablation.tex
\begin{figure}[ht]
    \centering 

    \begin{subfigure}{0.49\linewidth}
        \centering
        \includegraphics[width=\linewidth]{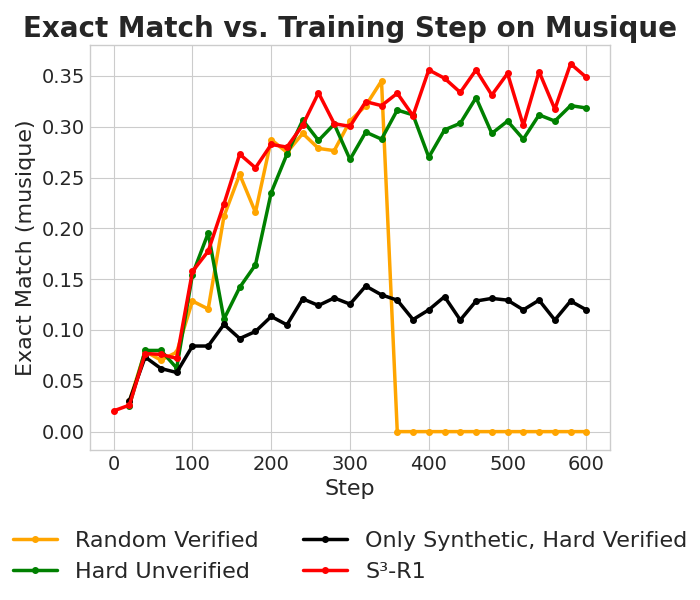}
        \subcaption{Exact Match (EM) scores during training.}
        \label{fig:rl_ablation_em}
    \end{subfigure}
    \hfill 
    \begin{subfigure}{0.49\linewidth}
        \centering
        \includegraphics[width=\linewidth]{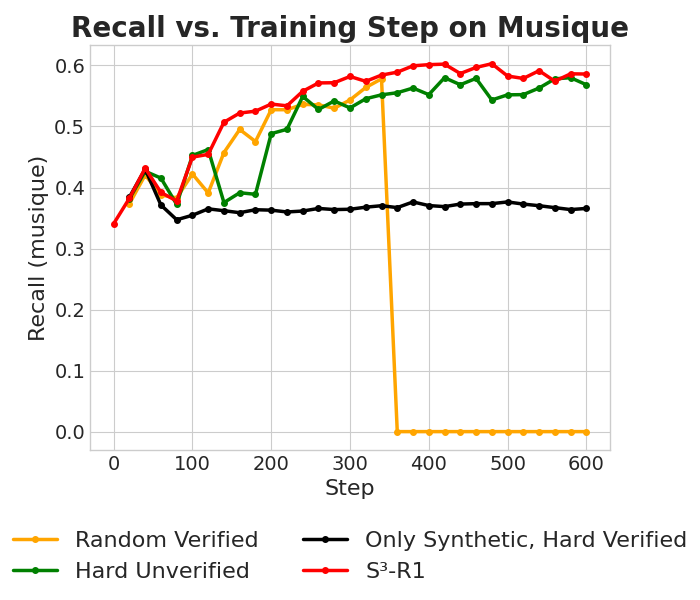} 
        \subcaption{Recall scores during training.}
        \label{fig:rl_ablation_recall}
    \end{subfigure}

    \caption{\textbf{Ablation of synthetic data generation components on training.} The left figure shows the Exact Match performance, while the right shows Recall. Our model trained with our RL enhancement on a mixture of original and synthetic data obtained from our proposed pipeline (Red) outperforms all other variants for compiling synthetic data on Pass@1 performance.}
    \label{fig:synth_ablation_combined} 
    \vspace{-1em}
\end{figure}

%% file: sections/conclusion.tex
In this work, we tackled two key bottlenecks in training LLMs for multi-hop search and QA, limited coverage of complex training questions, and sparse, outcome-only rewards. \methodname\ combines (i) a synthetic generation-and-curation pipeline that expands the training distribution with verified, intermediate-hardness multi-hop questions, and (ii) a retrieval-aware, denser RL learning signal that rewards not only final-answer correctness but also intermediate search quality and evidence selection. Together with stabilization techniques that prevent policy collapse during long-horizon optimization, these components yield more effective search-and-synthesis policies and robust generalization across both in-domain and out-of-domain evaluations on multi-hop QA datasets.

\section{Future Work}

Building on these results, a key direction is to further strengthen both the data pipeline and the supervision used for long-horizon search. On the data side, we will explore using stronger generative models to decompose complex multi-hop questions into sequences of grounded sub-questions, enabling a more structured curriculum and finer-grained control over question hardness during verification. On the optimization side, we plan to enrich our retrieval-aware reward with explicit process supervision, providing step-level feedback on query formulation, evidence selection, and stopping decisions. We expect that jointly scaling hardness-aware curricula and process-level learning signals will be central to training more robust long-horizon search agents.

%% file: sections/appendix.tex
\newpage
\section{Appendix}
\subsection{Training Details}

\paragraph{Datasets.}
We train exclusively on the MuSiQue training split, and evaluate using Exact Match (EM) on the test/validation sets of MuSiQue (in-domain) and CoFCA, 2WikiMultiHopQA, and HotpotQA (out-of-domain). Unlike Search-R1, which trains on a HotpotQA+NQ mixture, we deliberately use MuSiQue due to its smaller size (supporting faster, reproducible iterations) and its diverse composition of multi-step reasoning questions that align with our multi-hop objectives. 

\paragraph{Synthetic data mixture.}
To rebalance toward harder, multi-hop queries, we augment the training pool with verified synthetic questions produced by our pipeline (hard-seed mining, few-shot question generation with dissimilarity constraints, and oracle–vs–retrieval verification). Augmented models sample from the union of original and synthetic items during RL rollouts; unaugmented baselines use only the original corpus. Unless otherwise stated, retrieval and optimization settings are identical across augmented and unaugmented runs.

\paragraph{Models and retrieval.}
All experiments use the Qwen-2.5-7B Instruct model. We choose this model for experiments as they were shown to be the most performant in \cite{jin2025searchr1}. For retrieval, we index the entire corpus of documents referenced within the Musique dataset with the BM25 retriever. Each retrieval-based method consumes the top-5 passages per query.

\begin{table}[ht]
\centering
\caption{Training and rollout hyperparameters. Common settings apply to both Search-R1 and \methodname\ unless noted.}
\label{tab:exp_hparams}
\small
\begin{tabular}{p{0.42\linewidth}p{0.22\linewidth}p{0.22\linewidth}}
\toprule
\textbf{Hyperparameter} & \textbf{Search-R1} & \textbf{\methodname} \\
\midrule
Training steps & 600 & 600 \\
Policy LLM learning rate & 1e-6 & 1e-6 \\
Warm-up ratio (policy) & 0.285 & 0.285 \\
Samples per prompt (rollouts) & 5 & 5 \\
KL regularization coefficient $\beta$ & 0.001 & 0.001 \\
Entropy Coefficient $\lambda$ & 0.005 & 0.005 \\
Clip ratio Low $\epsilon_{\text{low}}$ & 0.2 & 0.2 \\
Clip ratio High $\epsilon_{\text{high}}$ & 0.2 & 0.32 \\
Negative Advantage Clip $\epsilon_{\text{neg}}$ & -- & 3.0 \\
Total / mini / micro batch size & 256 / 128 / 32 & 256 / 128 / 32 \\
Max input sequence length (tokens) & 8192 & 8192 \\
Max response length (tokens) & 500 & 500 \\
Max retrieved content (tokens) & 2048 & 2048 \\

Action budget $B$ (search steps) & 5 & 5 \\
Retriever top-$k$ (passages) & 5 & 5 \\
Rollout temperature / top-$p$ & 1.0 / 1.0 & 1.0 / 1.0 \\
vLLM tensor-parallel / mem util. & 1 / 0.6 & 1 / 0.6 \\
Checkpoint save frequency (steps) & 100 & 100 \\
Divergence handling & \multicolumn{2}{l}{Evaluate most recent stable checkpoint} \\
Compute and parallelism & \multicolumn{2}{l}{Single node, 8$\times$H100; FSDP with CPU offload} \\
\bottomrule
\end{tabular}
\end{table}

\subsection{Prompts}
This section details the prompts used for synthetic data generation, verification, and baseline evaluations.

\subsubsection{Synthetic Data Generation and Verification}

\paragraph{Question Generation.} This prompt is used to generate new, complex questions from a given set of documents.

\begin{tcolorbox}[colback=black!5!white,colframe=black!75!black,title=Question Generation Prompt]
You are an expert at creating high-quality questions from a given text corpus.
Your task is to generate a single, new, and insightful question that can be answered *only*
by using the information present in the provided documents. The new question should be
different from the original question provided. The question should require synthesizing
information, not just simple fact retrieval.

Now, based on the following documents, generate a new question.
The new question must be different from the original question.
First think carefully about the provided documents, and then generate a question based on them. Make sure you generate questions that would require all the documents to answer and not just a simple fact retrieval from the documents.
Provide only the question itself, without any preamble.
\end{tcolorbox}

\paragraph{Answer Verification.} This prompt is used to verify if a generated question can be answered from a set of ground-truth documents, ensuring data quality.

\begin{tcolorbox}[colback=black!5!white,colframe=black!75!black,title=Answer Verification Prompt]
You are a Question Answering system. Your task is to answer the following question based *only* on the information present in the provided documents. Do not use any external knowledge. Explicitly think about the question. Not all documents might be relevant to the question, hence you must carefully analyze which documents would be useful to answering your question. Finally provide an answer once you are reasonably confident of the answer.

**META INSTRUCTION:**
Provide the most concise or shortest answer as possible that is no longer than 5 words. If the answer cannot be found in the documents, respond with 'Answer not found in documents.'

--- DOCUMENTS ---
\texttt{{documents}}

--- QUESTION ---
\texttt{{question}}

--- ANSWER ---
\end{tcolorbox}

\subsubsection{Baseline Evaluation Prompts}

\paragraph{Standard RAG.} This prompt instructs the model to answer a question based only on a provided context, with a meta-instruction for conciseness.

\begin{tcolorbox}[colback=black!5!white,colframe=black!75!black,title=Standard RAG Prompt]
You are a Question Answering system. Your task is to answer the following question based *only* on the information present in the provided documents. Do not use any external knowledge. Explicitly think and reason about the question step by step. Not all documents might be relevant to the question, hence you must carefully analyze which documents would be useful to answering your question. Finally provide an answer once you are reasonably confident of the answer.

**META INSTRUCTION:**
Provide the most concise or shortest answer as possible that is no longer than 5 words. If the answer cannot be found in the documents, respond with 'Answer not found in documents'.

--- Information ---
\texttt{{search\_results}}

--- Question ---
\texttt{{question}}

Answer:
\end{tcolorbox}

\paragraph{Decomposition-based RAG.} This method uses two prompts: one to decompose the question and another to synthesize the final answer.

\begin{tcolorbox}[colback=black!5!white,colframe=black!75!black,title=Decomposition Prompt]
You are an advanced AI designed to function as a sophisticated query decomposer. Your primary objective is to meticulously analyze complex, multi-faceted questions and systematically break them down into a series of simpler, discrete, and self-contained sub-questions. Each sub-question should be formulated in such a way that it can be independently and effectively answered by a standard search engine, facilitating a step-by-step approach to resolving the original complex query.

Your output must adhere to a strict JSON format. Specifically, you are required to return the generated sub-questions as a JSON list of strings. This JSON list should be encapsulated within a JSON markdown block for clear presentation and easy parsing.

For illustrative purposes, consider the following example:
**Question:** What is the capital of the southernmost country in the continent with the longest river in the world?
**Sub-questions:**
\begin{verbatim}
[
  "What is the longest river in the world?",
  "Which continent has the longest river in the world?",
  "What is the southernmost country of that continent?",
  "What is the capital of that country?"
]
\end{verbatim}
Please provide the sub-questions for the following complex query:
**Question:** \texttt{{question}}
**Sub-questions:**
\end{tcolorbox}

\begin{tcolorbox}[colback=black!5!white,colframe=black!75!black,title=Synthesis Prompt]
You are an advanced Question Answering system designed to provide precise and concise answers. Your primary objective is to address the *Original Question* by leveraging information exclusively from the *Documents* and the insights gained from answering the *Sub-questions*. The *Sub-questions* are meticulously crafted, discrete, and self-contained inquiries derived from the *Original Question*, serving as a structured approach to facilitate a comprehensive answer. You must thoroughly analyze the *Original Question* and each *Sub-questions*, then synthesize the relevant information from the *Documents* to formulate your final response. It is imperative that you strategically utilize the *Sub-questions* to explicitly think step by step and guide your information extraction and synthesis from the *Documents*. Under no circumstances should you incorporate any external knowledge beyond the provided *Documents*.

**META INSTRUCTION:** Your answer must be the most concise possible, limited to a maximum of 5 words. If the definitive answer cannot be explicitly found in the provided *Documents*, you must respond with the exact phrase 'Answer not found in documents'.

--- Original Question ---
\texttt{{question}}

--- Sub-questions ---
\texttt{{sub\_questions\_formatted}}

--- Documents ---
\texttt{{aggregated\_context}}

Answer:
\end{tcolorbox}
\newpage
\paragraph{Iterative Multi-Hop RAG.} This prompt guides the model to perform a sequence of search-and-think steps, one hop at a time.

\begin{tcolorbox}[colback=black!5!white,colframe=black!75!black,title=Multi-Hop RAG Prompt]
You are a helpful assistant who will answer the given question step by step.
First, think about the question in \texttt{<think>}\texttt{</think>} tags.
Then, if needed, create a search query in \texttt{<search>}\texttt{</search>} tags.
Information will be provided to you in \texttt{<information>}\texttt{</information>} tags.
You can perform multiple searches.
When you have enough information, provide the final answer between \texttt{<answer>}\texttt{</answer>} tags.

Here is a correct example of a sample response every hop. Note that you are only allowed to think, search and answer once per hop. Information is provided to you based on the search query through an external tool.

**Question:** What is the capital of the southermost country of the continent with the longest river in the world?
**Hop 1:** \texttt{<think>} I need to know the longest river \texttt{</think>} \texttt{<search>} longest river \texttt{</search>}
**Hop 2:** \texttt{<think>} From infomation Nile is the longest river, I need to know continent \texttt{</think>} \texttt{<search>} continent where Nile is located \texttt{</search>}
**Hop 3:** \texttt{<think>} From information, Nile is located in Africa, lets find the southernmost country in Africa \texttt{</think>} \texttt{<search>} southernmost country in Africa \texttt{</search>}
**Hop 4:** \texttt{<think>} southermost country of Africa is South Africa, let search for its capital \texttt{</think>} \texttt{<search>} capital of South Africa \texttt{</search>}
**Hop 5:** \texttt{<think>} Based on the docs its Pretoria \texttt{</think>} \texttt{<answer>} Pretoria \texttt{</answer>}

**Bad examples:** Generating your own information per hop is not permitted.
Hop 1: \texttt{<think>} I need to know the longest river \texttt{</think>} \texttt{<search>} longest river \texttt{</search>} \texttt{<information>} ... \texttt{</information>}

**Bad examples:** Generating multiple hops in a single response is not permitted.
Hop 1: \texttt{<think>} I need to know the longest river \texttt{</think>} \texttt{<search>} longest river \texttt{</search>}
Hop 2: \texttt{<think>} From infomation Nile is the longest river, I need to know continent \texttt{</think>} \texttt{<search>} continent where Nile is located \texttt{</search>}

**META INSTRUCTION:**
Provide the most concise or shortest answer as possible that is no longer than 5 words. If the answer cannot be found in the documents, respond with 'Answer not found in documents'.

You must only search, think or answer once every hop.
You must not generate or use the \texttt{<information>} and \texttt{</information>} to hallucinate information, it will be provided to you.
Do not use any external knowledge when drafting the query, thinking step and answer. Use information you have gathered through search queries.

**Question:** \texttt{{question}}
\end{tcolorbox}

\newpage
\subsubsection{RL training Prompt}
For reinforcement learning, the model is guided by the following instruction-based prompt. This prompt encourages the generation of explicit reasoning and search actions, which form the basis of the policy's action space.

\begin{tcolorbox}[colback=black!5!white,colframe=black!75!black,title=RL Training Prompt]
You are a helpful assistant who will answer the given question with some potentially useful context step by step. 
You should analyze the question carefully, evaluate the given context (which may or may not be useful), and then generate an accurate and well-reasoned response. 
You should first have a reasoning process in mind. You must show your reasoning in \texttt{<think>} \texttt{</think>} tags. 
After reasoning, if you find you lack some knowledge, you create a query for a search engine enclosed within the \texttt{<search>} \texttt{</search>} tags and it will provide you the necessary information between the \texttt{<information>} and \texttt{</information>}. 
You must not generate or hallucinate any information in the \texttt{<information>} and \texttt{</information>} tags, it will be provided to you.
You can search as many times as you need. If you think you have sufficent information to answer, you must return final answer in \texttt{<answer>} \texttt{</answer>} tags.

Here are some examples for you to follow, note that the information is provided from the search engine, and you can ask for more information if you need. 
For example:
Question: What is the capital of China?
\texttt{<think>} I must search the capital of the China \texttt{</think>} \texttt{<search>} Capital of China \texttt{</search>} \texttt{<information>} Beijing is the capital of China \texttt{</information>} \texttt{<think>} Based on the information its Beijing\texttt{</think>} \texttt{<answer>} Beijing \texttt{</answer>}. 

Question: Is Beijing the capital of China?
\texttt{<think>} I must search the capital of the China \texttt{</think>} \texttt{<search>} Capital of China \texttt{</search>} \texttt{<information>} Beijing is the capital of China \texttt{</information>} \texttt{<think>} Based on the information its Beijing\texttt{</think>} \texttt{<answer>} Yes \texttt{</answer>}. 

Question: How many capitals did China have?
\texttt{<think>} The user is asking for capitals of China in the present and past \texttt{</think>} \texttt{<search>} List current capital of China \texttt{</search>} \texttt{<information>} Beijing is the capital of China \texttt{</information>} 
\texttt{<think>} I only know the present capital from the given information, I need to search for any capitals of China in the past\texttt{</think>} \texttt{<search>} List past capitals of China \texttt{</search>} \texttt{<information>} There are traditionally four major historical capitals of China referred to as the "Four Great Ancient Capitals of China". The four are Beijing, Nanjing, Luoyang and Xi'an (Chang'an). \texttt{</information>} \texttt{<think>} Based on the information its 4\texttt{</think>} \texttt{<answer>} 4 \texttt{</answer>}. 

Question: \texttt{{question}}
\end{tcolorbox}

\section{LLM Usage Declaration}

We used Gemini 2.5 Pro to improve the grammar and flow of the final draft. We also used LLMs to paraphrase and condense the verification section in the Methods, based on a human-written version. Additionally, we used the Overleaf AI Assistant to address formatting issues in the paper’s tables and figures. Figure \ref{fig:pipeline} was generated with assistance from Nano Banana Pro based on a flowchart created by the authors on slides.